\documentclass[runningheads]{llncs}
\usepackage[T1]{fontenc}
\usepackage{graphicx}
\usepackage{url}
\begin{document}
\title{Automatic Registration of SHG and H$\&$E Images with Feature-based Initial Alignment and Intensity-based Instance Optimization: Contribution to the COMULIS Challenge}
\titlerunning{Automatic Registration of SHG and H\&E Images}
\author{Marek Wodzinski\inst{1, 2}\orcidID{0000-0002-8076-6246} \and
Henning M\"{u}ller\inst{2, 3}\orcidID{0000-0001-6800-9878}}
\authorrunning{M. Wodzinski and H.  M\"{u}ller}
%
\institute{
$^{1}$AGH University of Krakow, Department of Measurement and Electronics \\ Krakow, Poland \\
$^{2}$University of Applied Sciences Western Switzerland (HES-SO Valais-Wallis) \\ Information Systems Institute, Sierre, Switzerland \\
$^{3}$ University of Geneva, Medical Faculty, Geneva, Switzerland \\
\email{
wodzinski@agh.edu.pl \\
}
}
\maketitle              
\begin{abstract}
The automatic registration of noninvasive second-harmonic generation microscopy to hematoxylin and eosin slides is a highly desired, yet still unsolved problem. The task is challenging because the second-harmonic images contain only partial information, in contrast to the stained H\&E slides that provide more information about the tissue morphology. Moreover, both imaging methods have different intensity distributions. Therefore, the task can be formulated as a multi-modal registration problem with missing data. In this work, we propose a method based on automatic keypoint matching followed by deformable registration based on instance optimization. The method does not require any training and is evaluated using the dataset provided in the Learn2Reg challenge by the COMULIS organization. The method achieved relatively good generalizability resulting in 88\% of success rate in the initial alignment and average target registration error equal to 2.48 on the external validation set. We openly release the source code and incorporate it in the DeeperHistReg image registration framework.

\end{abstract}
\section{Introduction}

The automatic registration of microscopy images is still an active research area~\cite{ANHIR1,ACROBAT1}. Recently, several notable contributions were proposed to automatic registration of whole slide images (WSIs) acquired using different stains, varying from contributions that focus on the quality of deformable registration~\cite{zhao2019unsupervised,wodzinski2021deephistreg,ge2022unsupervised,wodzinski2021multistep}, through methods that address the robustness of initial alignment~\cite{marzahl2021robust,lin2023end,pyatov2022affine,wodzinski2024regwsi}, ending with methods that propose ready-to-use software packages that can perform automatic registration without time-consuming parameter tuning or deep network retraining~\cite{VALIS,deeperhistreg,HyReCo1}.

\begin{figure*}[!htb]
    \centering
    \includegraphics[width = 0.95\textwidth]{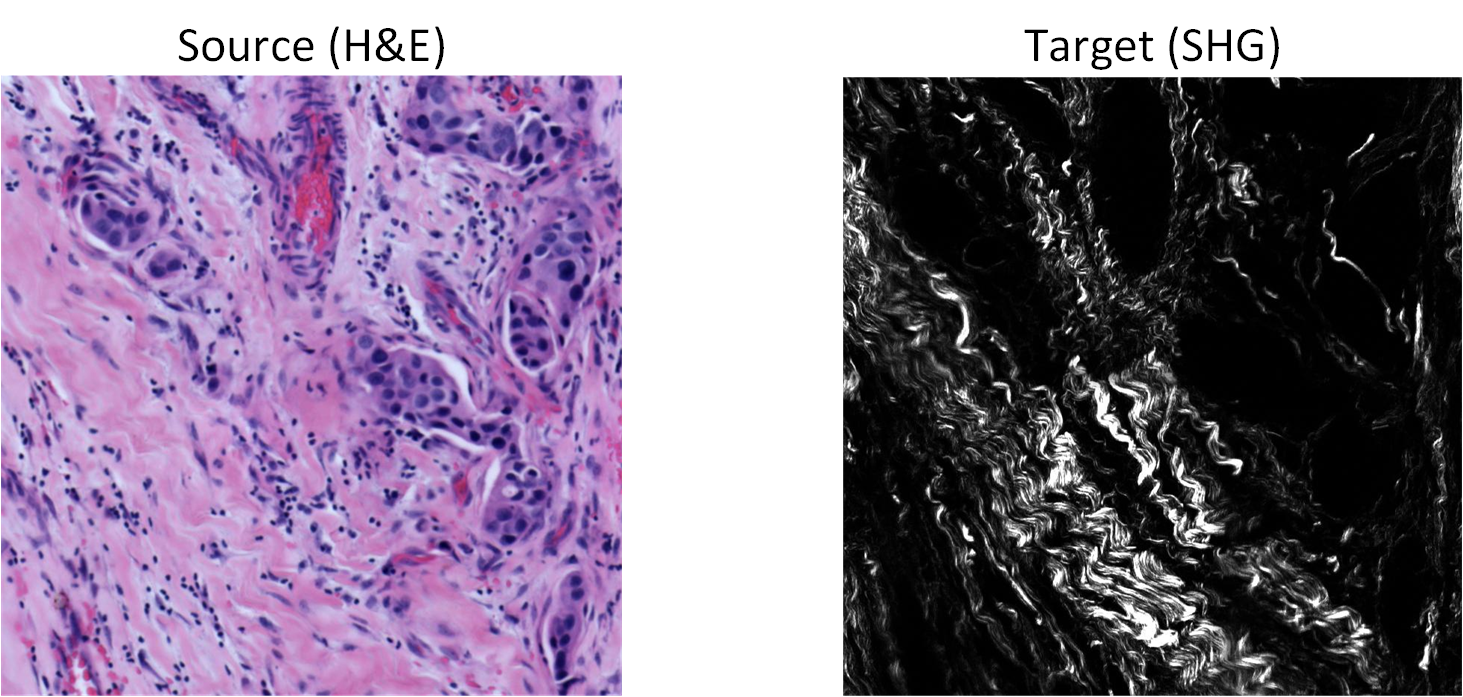}
    \caption{Exemplary pair of H\&E and SHG images. Note the significantly different intensity distributions and the amount of missing data in the SHG image.}
    \label{fig:example}
\end{figure*}

Nevertheless, the registration of microscopic images acquired using different imaging modalities is still an unsolved problem that requires further investigation. An example of a highly important task within the research area is the automatic registration of second-harmonic generation (SHG) microscopy images to bright-field images stained by hematoxylin and eosin (H\&E) or immunohistochemistry (IHC) dyes. SHG microscopy is a non-invasive imaging technique, does not require exogenous labels, and allows one to easily visualize collagen fibers in the tissue microenvironment. This enables one to perform real-time observations without introducing the toxicity associated with staining agents or external artifacts. However, it only provides partial information while H\&E or IHC present more information about the tissue morphology. Moreover, both imaging modalities have significantly different intensity distributions. Therefore, a dedicated solution that addresses both problems is required. The importance of the task motivated researchers from the COMULIS consortium to organize a dedicated sub-challenge during the well-recognized Learn2Reg challenge~\cite{hering2022learn2reg}. An exemplary pair of SHG and H\&E images is shown in Figure~\ref{fig:example}.

So far, the work on SHG and H\&E was rather limited. The researchers focused primarily on the registration of H\&E to IHC slides~\cite{ANHIR1,ACROBAT1} or the H\&E to magnetic resonance images (MRIs)~\cite{shao2021prosregnet,alyami2022histological,alegro2016multimodal}. There are several different methods dedicated to the initial alignment, including both intensity-based methods~\cite{lin2023end,awan2023deep,pyatov2022tahir}, as well as the feature-based contributions~\cite{pyatov2022affine}. The following deformable registration is still usually solved by iterative optimization due to the tremendous heterogeneity of WSIs~\cite{HyReCo1,wodzinski2024regwsi,VALIS,wodzinski2021multistep,lotz2015patch}, however several deep learning-based methods exist~\cite{zhao2019unsupervised,wodzinski2021deephistreg,ge2022unsupervised}, unfortunately without exhaustive evaluation of the generalizability to previously unseen distributions.  Methods often struggle with the following challenges related to the registration of WSIs: (i) large initial misalignment, (ii) tremendous size of input images, and (iii) missing data caused by different properties of staining agents. Nevertheless, there is an influential work presenting an automatic intensity-based method for the initial alignment of SHG and H\&E images~\cite{keikhosravi2020intensity}. The method is based on an initial preprocessing that decreases the amount of information in the bright field images, followed by an iterative optimization of the initial alignment matrix. Even though the method is successful, it may struggle with recovering large rotational initial misalignment.

\textbf{Contribution: } In this work, we propose an automatic registration method dedicated to SHG and H\&E images. The method consists of deep feature-based initial alignment allowing recovery of large deformations, followed by an optional deformable registration based on the instance optimization. We evaluate the method using the dataset released by the COMULIS organization during the Learn2Reg challenge. We show considerable generalizability and quality of the proposed initial alignment and argue whether the deformable registration is really necessary. We openly release the source code and incorporate the method in one of the open-source registration packages dedicated to WSIs, enabling the users to directly use it in their research.

\section{Method}

\subsection{Overview}

\begin{figure*}[!htb]
    \centering
    \includegraphics[width = 0.7\textwidth]{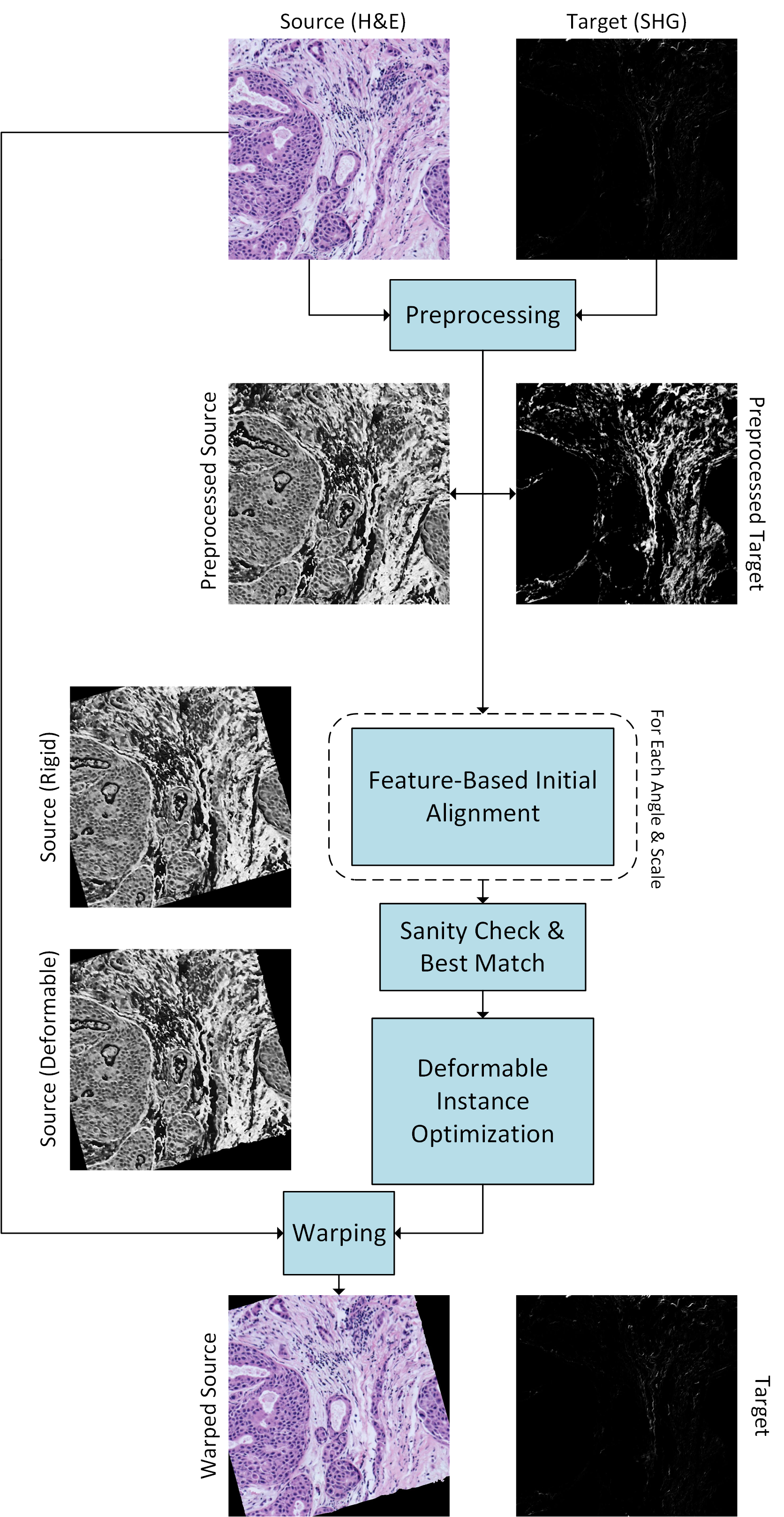}
    \caption{Visualization of the registration pipeline presenting also the intermediate results. Best viewed zoomed in. The target SHG image is presented in the original intensity range to emphasize the task difficulty (zoom required to see the details).}
    \label{fig:pipeline}
\end{figure*}

The proposed method consists of three steps: (i) preprocessing to make the features from both modalities as similar as possible, (ii) feature-based sparse initial alignment based on SuperPoint and SuperGlue, and (iii) dense deformable registration based on the instance optimization. The processing pipeline is presented in Figure~\ref{fig:pipeline}.

\subsection{Preprocessing}

Since the SHG and H\&E images have substantially different intensity distributions and visualize distinct structures it is crucial to perform preprocessing that makes the general geometric features as similar as possible. 

The H\&E images are firstly converted to the HSV color space. Then, only the hue image is selected, normalized to [0-1] intensity range, and equalized by global histogram equalization. Then the image is filtered by a 5x5 median filter. The preprocessing of SHG images is similar with the difference that the images are equalized and filtered directly, without the conversion to HSV color space. The registration is performed using the original resolution of the images, however, for the feature-based initial alignment other resolution level are used as well (100, 200, 300, 400, and 500 pixels in each dimension).

\begin{figure*}[!htb]
    \centering
    \includegraphics[width = 0.95\textwidth]{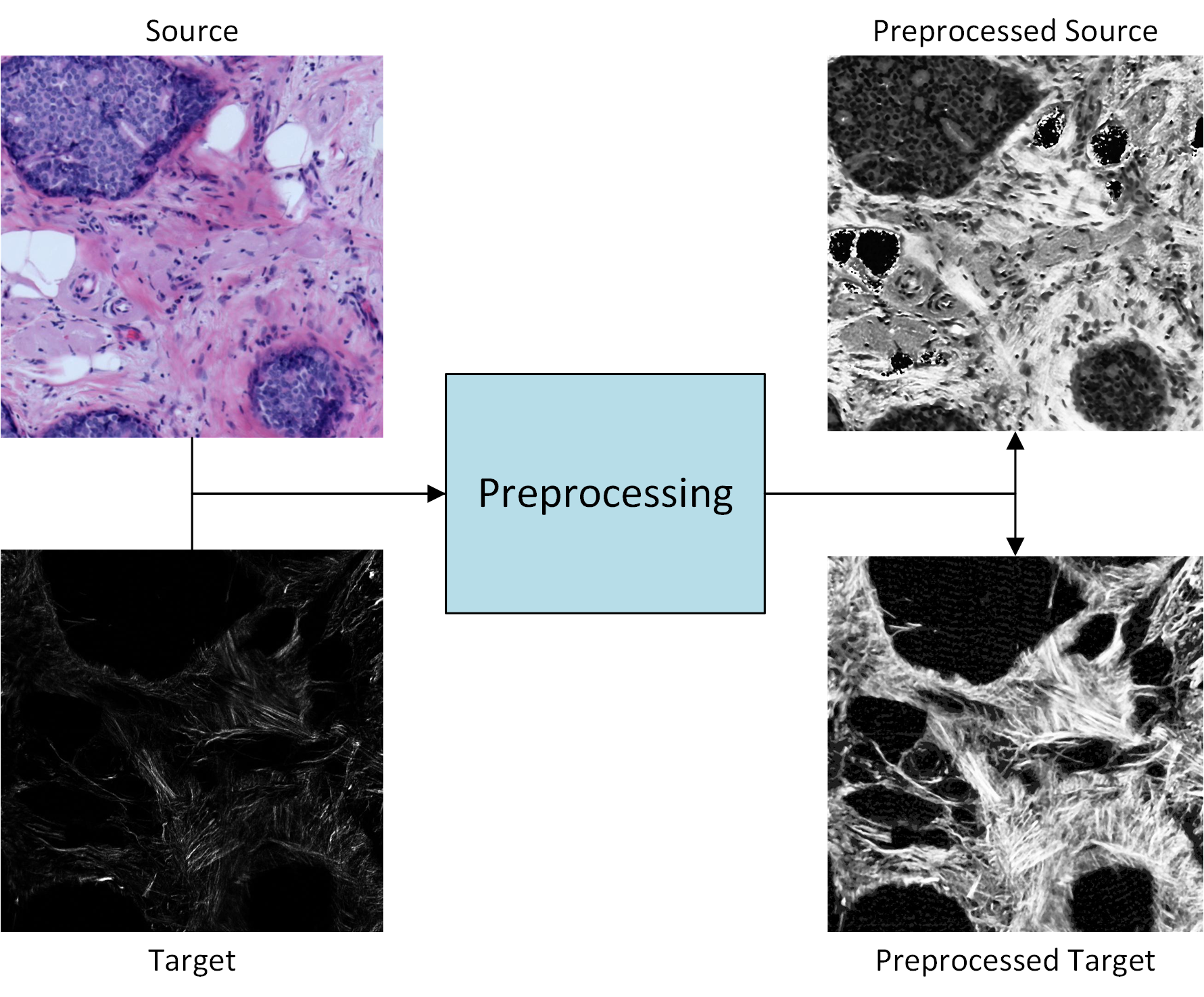}
    \caption{Visualization of the proposed preprocessing.}
    \label{fig:preprocessing}
\end{figure*}

The reason why we decided to perform the histogram equalization is connected with the fact that the distributions of the hue channel of H\&E and the raw SHG images have significant outliers. Such distributions decrease the robustness of the following feature-based alignment. The median filtering was implemented because the images are strongly influenced by the impulse noise. An exemplary pair of images before and after the preprocessing is shown in Figure~\ref{fig:preprocessing}.

\subsection{Initial Alignment}

The initial registration is implemented as a feature-based alignment of sparse key points using the SuperPoint keypoint extractor~\cite{detone2018superpoint} and SuperGlue matcher~\cite{sarlin2020superglue}. They can be considered as foundation models dedicated to learning-based feature matching. Our previous works confirmed their superior generalizability into previously unseen cases and intensity distributions~\cite{wodzinski2024regwsi}. 

The SuperPoint and SuperGlue are not rotation and scale invariant. Therefore, we incorporate them in an exhaustive search that calculates the transformation and number of matches for each predefined resolution and initial rotation. Then, we filter the transformations that do not meet the predefined scale criteria (increase or decrease in scale more than 10\%) and finally return the transform associated with the highest number of matched key points.

Finally, we compare the SuperPoint/SuperGlue pair to the classical SIFT + RANSAC method~\cite{lowe2004distinctive} and the recent OmniGlue model~\cite{jiang2024omniglue} showing its superiority. All the methods are evaluated in the same exhaustive framework.

\subsection{Deformable Registration }

The deformable registration is implemented as a multilevel instance optimization, an iterative optimization process. We decided to avoid training a deep network dedicated to the deformable registration because, with the available number of training samples, the method would not generalize into previously unseen cases, especially considering the fact that there are not openly available foundation models dedicated to deformable microscopy registration that could be used for fine-tuning the network in low data regimes.

The objective function is defined as a weighted sum of the local mutual information (MI) as the similarity metric, and the diffusive regularization as the regularizer, defined as:
\begin{equation}
    O_{REG}(S, T, u) = MI(S_i \circ u_i, T_i) + \theta_i Reg(u_i),
\end{equation}
where $S_i, T_i$ are the source and target images at the i-th resolution level respectively, $u_i$ is the calculated displacement field, $\theta_i$ denotes the regularization coefficient at i-th resolution level, $MI$ denotes the local version of mutual information, $Reg$ is the diffusive regularization, $\circ$ denotes the warping operation, and $N$ is the number of resolution levels.

\subsection{Dataset \& Experimental Setup}

The dataset consists of 166 pairs of SHD and H\&E images among which 156 represent an unannotated training set and 10 are validation pairs with ground truth accessible through the Grand-Challenge platform. The images present human breast and pancreatic cancer tissue and were acquired at 40x magnification~\cite{keikhosravi2020intensity,eliceiri2020multimodal} at the Laboratory for Optical and Computational Instrumentation, Department of Biomedical Engineering, University of Wisconsin-Madison, Madison, WI 53706, USA. The dataset was released under a CC license and is used as a subtask in the Learn2Reg 2024 challenge.

All experiments were performed using a workstation equipped with a NVIDIA RTX A6000 GPU. We performed ablation studies to verify the robustness of different initial alignment methods. The source code and the hyperparameters used are openly available and incorporated in the \textit{DeeperHistReg} registration framework~\cite{deeperhistreg}.

\section{Results}

We evaluate the proposed approach using three methods: (i) target registration error (TRE) reported using the validation set with ground truth available through the Grand-Challenge platform, (ii) the ratio of successful initial alignments (evaluated both on the validation and the training set), (iii) visual analysis of the registered images. We also used the training set for the evaluation because the proposed method was not re-trained or fine-tuned using the samples available in the training set.

\begin{figure*}[!htb]
    \centering
    \includegraphics[width = 0.95\textwidth]{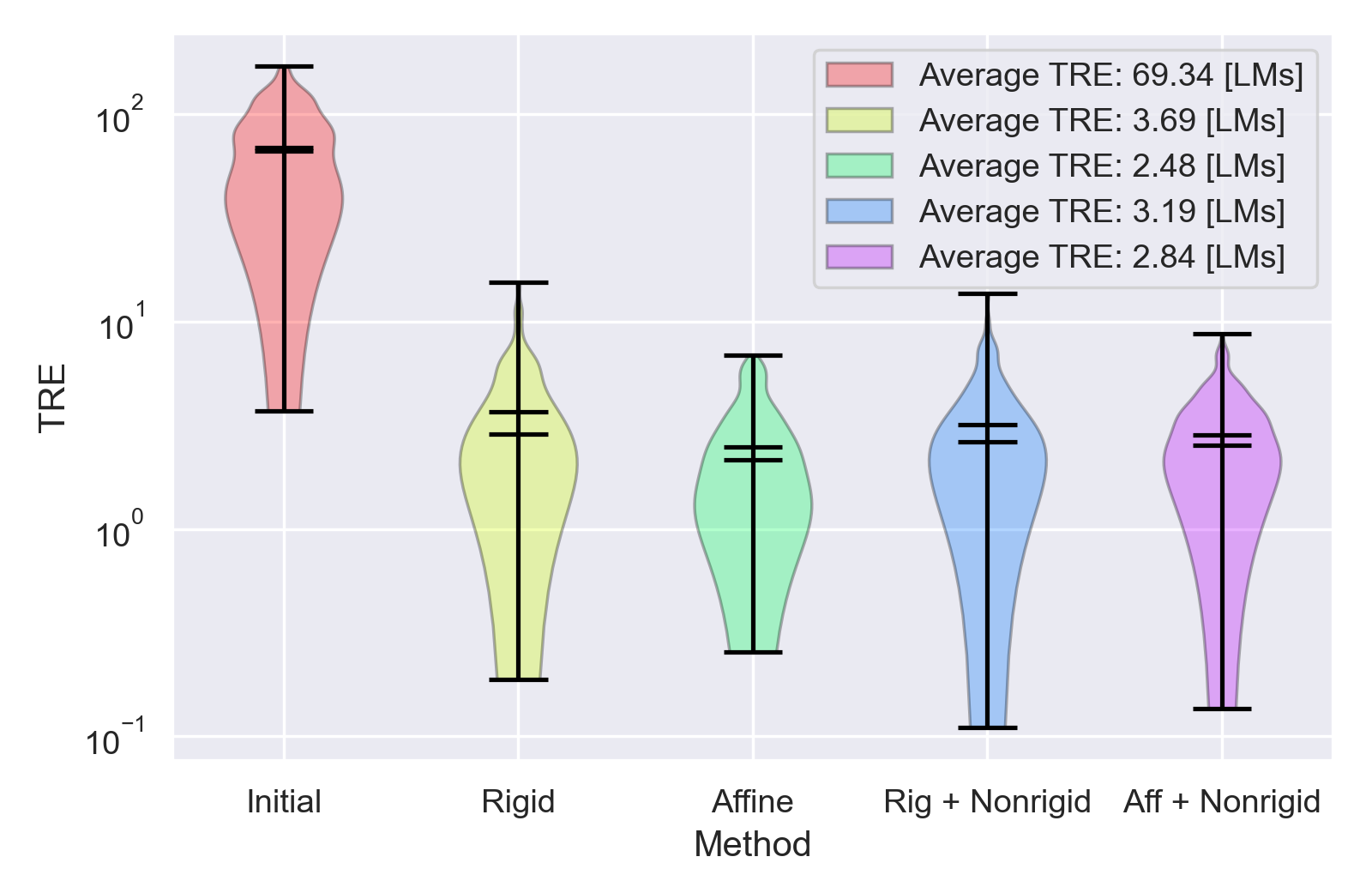}
    \caption{The TRE calculated for the validation pairs using the Grand-Challenge platform.}
    \label{fig:tre}
\end{figure*}

\begin{figure*}[!htb]
    \centering
    \includegraphics[width = 0.92\textwidth]{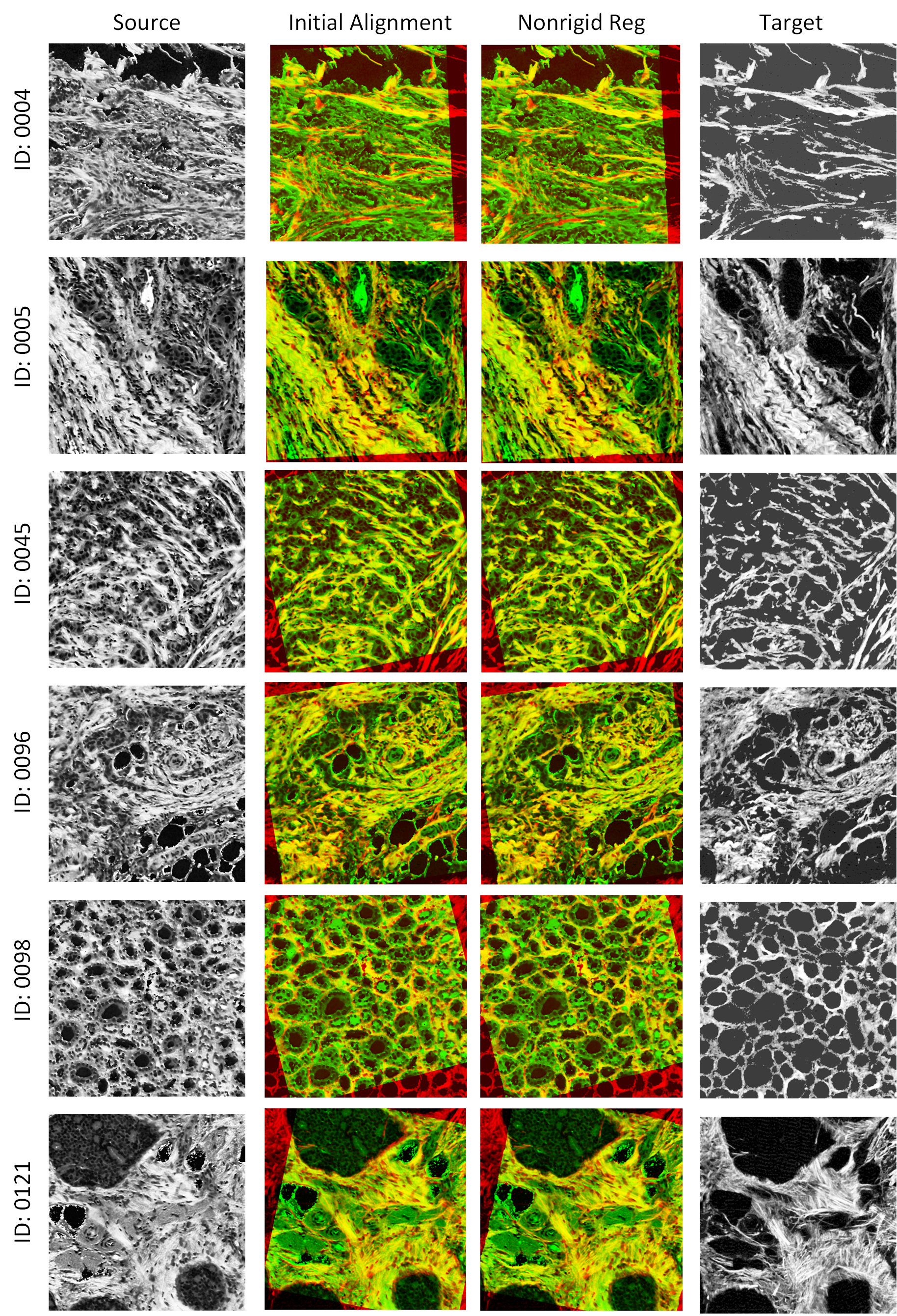}
    \caption{Qualitative registration results using several samples from the validation subset. The source and target images are overlayed in different color channels to present the alignment quality. The preprocessed images are used for the presentation clarity. Note the small impact of the deformable registration.}
    \label{fig:visual_results}
\end{figure*}

The TRE is presented in Figure~\ref{fig:tre}. Due to a limited number of available submissions, we report only results presenting the influence of initial alignment and the following deformable registration. The percentage of successful initial alignments is presented in Table~\ref{tab:success_rate} and includes an ablation study related to the methods used in the feature-based registration. The success rate is the ratio of correctly registered pairs (visual inspection whether the images are meaningfully aligned) to obviously incorrect registrations (no improvement or worsening the results). To evaluate the success rate, we combined pairs from the training and validation subsets. Exemplary visualizations of the registration results are presented in Figure~\ref{fig:visual_results}.

\begin{table*}[!htb]
\centering
\caption{The success rate of several initial alignment algorithms. Note the generalizability of the SuperPoint and SuperGlue methods. All the methods (except the intensity-based iterative affine registration using mutual information) were incorporated in the multi-scale and multi-angle exhaustive search.}
\renewcommand{\arraystretch}{1.0}
\footnotesize
\resizebox{0.99\textwidth}{!}{%
\begin{tabular}{lcc}
\label{tab:success_rate}
Method & Success Rate All [\%] $\uparrow$ & Success Rate Val [\%] $\uparrow$ \tabularnewline \hline

SuperPoint \& SuperGlue & \textbf{87.95} & \textbf{100.00} \tabularnewline \hline
OmniGlue & 48.79 & 60.00 \tabularnewline \hline
SIFT + RANSAC & 34.34 & 30.00 \tabularnewline \hline
Iterative Multilevel Affine (MI) & 37.95 & 40.00 \tabularnewline \hline

\hline
\end{tabular}}
\end{table*}

\section{Discussion}

\begin{figure*}[!htb]
    \centering
    \includegraphics[width = 0.7\textwidth]{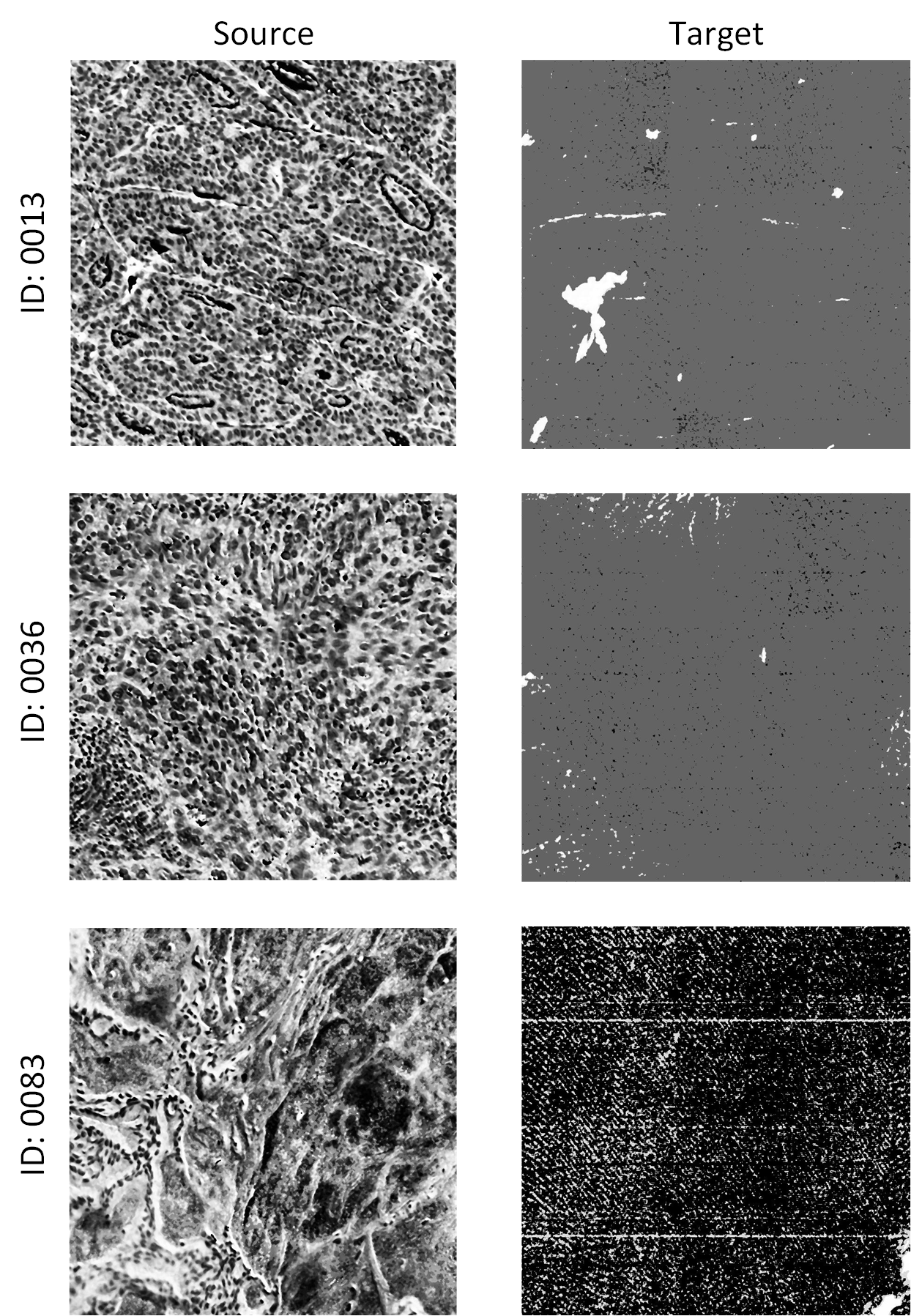}
    \caption{Examples of several incorrectly registered pairs, however, according to the authors opinion, such pairs cannot be registered correctly. The images are presented after the preprocessing for the presentation clarity.}
    \label{fig:impossible}
\end{figure*}

The proposed method achieved considerable robustness and generalizability (Figure~\ref{fig:tre}). The initial alignment method successfully registered all validation pairs and 88\% of all the registration pairs combined together (Table~\ref{tab:success_rate}). This is a significant achievement because: (i) part of the training pairs seem to be impossible to register (Figure~\ref{fig:impossible}), (ii) the images are so different that it is difficult to even capture the relations by humans. On the other hand, the following deformable registration should be considered as an optional step. It does not improve the registration quality for the majority of the slides (no statistically significant differences, Wilcoxon signed-rank test, p-value > 0.05), however, may introduce undesired foldings (even though the proposed method does not introduce any foldings). It seems that the samples were not undergoing large nonrigid deformations.

The proposed method has several limitations. The main one is connected with the initial alignment time. Since the combination of SuperPoint keypoint extractor and SuperGlue matcher is not scale and rotation invariant, it is required to perform the matching using various initial rotations and scales. As a result, the initial alignment requires an average of 40s to be computed (NVIDIA A6000). In fact, even methods that claim to be rotation and scale invariant (e.g. SIFT) express the property only in limited range of values. Thus, the exhaustive search of initial rotation and scale is beneficial for the robustness of the feature-based initial alignment. The second limitation is connected with the deformable registration. The images not only come from different intensity distributions but also present different structures (e.g. collagen fibers vs nuclei). This results in missing data that cannot be simply addressed by mutual information or other similarity metrics dedicated to multimodal registration. It is an open question whether the dense deformable registration is the correct approach to the problem, especially considering that the TRE is based on sparse key points that can be annotated only in regions with direct correspondences, thus the registration quality in other regions remains unknown.

Interestingly, the combination of SuperPoint and SuperGlue outperforms to more recent OmniGlue method. The observation is counterintuitive because the OmniGlue article~\cite{jiang2024omniglue} states a superior generalizability of the keypoint matcher. The claim is not supported by the results, and we show that, at least for multimodal microscopy images, the combination of SuperPoint and SuperGlue is superior to OmniGlue.

Finally, we point out an important observation related to the initial alignment and the available data. It would be significantly easier to register images for entire slides instead of cropped patches. For some of the cropped patches, it is inherently difficult to find the sparse correspondences, however, larger contextual information could result in a significantly higher robustness of the feature-based registration. Maybe in future editions of the challenge the task could be split into two problems: (i) an initial alignment using resampled whole slide images, and (ii) the deformable registration of the cropped patches.

To conclude, we proposed an automatic registration method dedicated to SHD and H\&E images. The method consists of dedicated preprocessing, robust feature-based initial alignment, and optional deformable registration. The method achieved considerable registration accuracy and robustness and is openly available to the scientific community.

\subsubsection{Acknowledgements} We gratefully acknowledge Polish HPC infrastructure PLGrid support within computational grant no. PLG/2024/017079. The research was partially supported by the program "Excellence Initiative - Research University" for AGH University.

\subsubsection{Disclosure of Interests} The authors have no competing interests to declare that are relevant to the content of this article.

%
%
\bibliographystyle{splncs04}
\bibliography{bibliography}
\end{document}